\documentclass[letterpaper, 10 pt, conference]{ieeeconf}  

\IEEEoverridecommandlockouts                              

\usepackage{amsmath} 
\usepackage{amssymb}  
\usepackage{graphicx}
\usepackage{verbatimbox}
\usepackage{subfig}

\title{\LARGE \bf
Maximum Consensus Localization using an Objective Function based on Helmert’s Point Error}

\author{Jeldrik Axmann$^{1}$, Yimin Zhang$^{2}$ and Claus Brenner$^{3}$
\thanks{$^{1}$Jeldrik Axmann is with the Institute of Cartography and Geoinformatics, Leibniz University Hannover, Germany
        {\tt\small jeldrik.axmann@ikg.uni-hannover.de}}%
\thanks{$^{2}$Yimin Zhang is with the Leibniz University Hannover, Germany
        {\tt\small yimin.zhang@stud.uni-hannover.de}}%
\thanks{$^{2}$Claus Brenner is with the Institute of Cartography and Geoinformatics, Leibniz University Hannover, Germany
        {\tt\small claus.brenner@ikg.uni-hannover.de}}%
}

\begin{document}
\maketitle
\thispagestyle{empty}
\pagestyle{empty}

\begin{abstract}

Ego-localization is a crucial task for autonomous vehicles. On the one hand, it needs to be very accurate, and on the other hand, very robust to provide reliable \textit{pose} (position and orientation) information, even in challenging environments. 
Finding the best ego-position is usually tied to optimizing an objective function based on the sensor measurements.
The most common approach is to maximize the likelihood, which leads under the assumption of normally distributed random variables to the well-known least squares minimization, often used in conjunction with recursive estimation, e.~g. using a Kalman filter.
However, least squares minimization is inherently sensitive to outliers, and consequently, more robust loss functions, such as $L_1$ norm or Huber loss have been proposed.
Arguably the most robust loss function is the outlier count, also known as {\em maximum consensus} optimization, where the outcome is independent of the outlier magnitude.

In this paper, we investigate in detail the performance of maximum consensus localization based on LiDAR data. We elaborate on its shortcomings and propose a novel objective function based on Helmert's point error.
%
In an experiment using 3001 measurement epochs, we show that the maximum consensus localization based on the introduced objective function provides superior results with respect to robustness.
\end{abstract}

\section{INTRODUCTION}

Arguably the most common approach for localization tasks is the fusion of measurements from different sensors using a Kalman filter \cite{Chen2012}.
By doing so, the localization solution is found based on a recursive least-squares minimization of measurement residuals \cite{Kalman1960}. 
When LiDAR data is used, these residuals are often computed based on the distance between measured and reference points or surfaces. Their minimization is also encountered in the context of point cloud registration.
There, the most common approach is the \textit{Iterative Closest Point} (ICP) algorithm \cite{Besl1992}, which iteratively alternates between the two steps of establishing correspondences (i.e., identifying inliers) and minimizing errors. 
Especially with respect to localization based on LiDAR data, outliers arise frequently due to erroneous measurements,
outdated maps, vegetation, and moving or temporary objects, like parked cars. 

To minimize the influence of outliers, the
maximum consensus criterion can be used. For this estimation strategy, the consensus set size is used as robust objective function and the candidate model yielding the largest consensus set and consequently the highest consistency with the postulated functional model is considered as solution. To prevent outliers from influencing the estimate, correspondences with residuals larger than a predefined threshold are not taken into consideration, 
which makes maximum consensus techniques inherently robust \cite{ChinSuter2017}.
Unfortunately, no efficient optimization strategy is available.
The most famous approach to tackle this is \textit{Random Sample Consensus} (RANSAC) \cite{Fischler1981}, which explores parameter space by randomly drawing samples.
This technique works well in practice and numerous variants have been proposed. However, being based on random draws, it does not guarantee to yield the globally optimal solution.

Axmann and Brenner \cite{Axmann2021} proposed a localization based on the maximum consensus criterion.
Point clouds from a Velodyne VLP-16 scanner (henceforth referred to as \textit{car sensor scans}) are aligned to a highly accurate, globally georeferenced point cloud (henceforth referred to as \textit{map}) to determine the vehicle poses.
If the distance between a car sensor scan point and a map point is smaller than a given threshold, this is counted as a match. The total count of those matches is used as objective function and the transformation yielding the highest consensus is determined by exhaustive search. This yields the global optimum.
While exhaustive search has a cost which is exponential in the number of search space dimensions, it is simply a (large) constant when the number of dimensions is fixed, as in our case. Exhaustive computation has the benefit of revealing the full distribution of the objective function, which is helpful during evaluation. In practical terms, the count operations are simple in nature and lead to {\em stencil} type operations well suited for parallel processing. In addition, computations can be sped up by using discrete optimization techniques, such as Branch-and-Bound, when transitioning to real-time applications.

%
While counting is a simple, fast and robust operation, it fails to model the exact role of correspondences for the determination of the unknown pose. In general, most correspondences are between LiDAR points and map (world) surfaces, for example, points located on facades, walls, road, and walkway surfaces. While each such ``hit'' should increase the confidence for the corresponding pose, it does not constrain the pose with respect to three dimensions. Rather, the correspondence imposes only one constraint, orthogonal to the surface.
In urban canyons, it is frequently the case that almost all matches stem from the ground and the parallel facades to the left and right of the vehicle, with only a few matches (e.~g.\ on balconies and other protruding facade parts) available to constrain the pose in longitudinal (driving) direction. In such cases, the weak constraints in the longitudinal direction are often outvoted by other effects, for example, fluctuating point densities along the facades. To counteract this, instead of simple counting, the actual constraint generated by each match must be modeled.
%
%

In this paper, we propose a novel robust objective function for point cloud registration, which incorporates the normal vectors of the matched map points. 
Due to the higher and more regular point density of the map point cloud, its normal vectors are more stable than the normal vectors of the car sensor scans. 
In particular, including all matches from the consensus set, a point-to-plane adjustment, which makes use of the normal vectors, is carried out. The covariance matrix of the estimated pose parameters will then give the errors with respect to the pose parameters, and its trace will be small if the errors are small and homogeneous in all directions.
In terms of the position component of the pose, the square root of the trace is also known as Helmert's point error. For our purposes, we define the inverse of the trace as the score. This makes sense because for $n$ i.~i.~d.\ measurements, the variance of an estimator will be proportional to $1/n$, so that the inverse of the trace can be seen as a generalization of the observation count.
%

We show that the new objective function solves the shortcomings of the original, count based maximum consensus localization.
In detail, (i)~we explain the localization based on the registration of sparse car sensor scans from a Velodyne VLP-16 to a dense high resolution map point cloud acquired by a mobile mapping system, (ii)~we introduce the new maximum consensus approach based on Helmert's point error, (iii)~we compare it to the original maximum consensus localization, and (iv)~we evaluate it on 3001 epochs from an inner city measurement drive. 

\section{Related Work}


\subsection{LiDAR based localization}
Localization approaches based on LiDAR data can be divided in three categories \cite{Elhousni2020}: Registration based, feature based, and deep learning based methods.
{\em Registration based approaches} refer to algorithms which are using the scan points directly.
The most common algorithm is the ICP, including its wide range of variants, especially the "point-to-plane" ICP \cite{Chen1992,Zhang1994} and the generalized ICP \cite{Segal2009}, which can be understood as "plane-to-plane" ICP.
To reduce the sensitivity towards outliers, enhancements such as
selection, matching or weighting criteria for finding correspondences or minimizing residuals are commonly used \cite{Chetverikov2002}.
Moreover, to cope with the large number of outliers in LiDAR data, preprocessing strategies especially focusing on outlier rejection have been introduced \cite{Bustos2018}.
In contrast to ICP frameworks, Chen et al. \cite{Chen2021} generate range images from real 3D LiDAR scans and synthetic range images from a mesh-based map, which are integrated in a Monte Carlo localization framework. For each particle representing a transformation, a synthetic range image is generated.  
To determine the resampling weight of each particle, its synthetic range image is compared to the measured range image with the mean absolute pixel-wise difference as the similarity measure.

In contrast to the direct registration of point clouds, {\em feature based approaches} only consider a relatively small number of extracted 3D spatial or intensity features. In 
\cite{Zhang2022} 
and \cite{Wei2020}, localization approaches based on curb and intensity 
features are proposed.
Schlichting and Brenner \cite{Schlichting2014} match poles and planes to a landmark map to determine the vehicle pose. For the assignment of measured and reference features, instead of using the nearest neighbor approach, feature patterns describing the spatial relation are proposed. Through the matching of feature patterns, the rate of false assignments is strongly reduced.
In \cite{Steinke2021}, pole, wall, and corner features are extracted and their spatial relation described as so-called `fingerprints'. Based on freely available street data, a reference feature map is established. For the localization, measured and map fingerprints are matched. The uniqueness of local pole patterns has been shown earlier by \cite{brenner2009}.
In \cite{Wolcott2014}, monocular images are registered against synthetically generated ground images from a LiDAR map. For a predefined set of heading angles, synthetic images are predicted and an exhaustive search is conducted to find the highest consistency between the real and a synthetic image indicating $xy$-translation and heading angle. 
Furthermore, graph-based approaches for multi sensor fusion exist. In \cite{Merfels2016}, LiDAR scans are matched to a globally georeferenced point cloud and the obtained pose information is fused with information from GPS, wheel odometry and a visual localization using a sliding window pose graph. 
In \cite{Wilbers2019}, pole-like landmarks are detected in LiDAR scans and matched against a map. The resulting pose information are optimized in a sliding window factor graph with the ability to correct map associations.

Finally, recent publications indicate an increasing interest in solving LiDAR based localization and odometry tasks using {\em deep learning}. In \cite{Elbaz2017}, an unsupervised machine learning approach for point cloud registration is proposed, which is based on super-points described by a low-dimensional geometric descriptor. 
In \cite{Tim2018} and \cite{Lu2019},
corrections between a traditional localization solution and ground-truth poses are learned. 
In \cite{Li2019}, real time LiDAR odometry including a map based refinement, in \cite{Wang2019}, LiDAR odometry by representing two consecutive scans as panoramic depth images, and in \cite{Cho2020}, LiDAR odometry based on an unsupervised learning approach with no need for ground truth labels is proposed. 
In \cite{Yin2018}, a siamese network for global localization, and in \cite{Lu2019a}, a 3D point cloud registration framework much less susceptible to inaccurate initialization due to the selection of robust and suitable key points are introduced.
For all these methods, but especially for deep learning, a key question is how to quantify and guarantee accuracy and reliability.

\subsection{Maximum consensus}
Maximum consensus approaches can be categorized into heuristic, deterministic and exact techniques. Heuristic maximum consensus strategies do not guarantee a globally optimal solution, however, they are usually able to provide good results in a reasonable amount of time. The arguably most famous heuristic approach is the already mentioned \textit{Random Sample Consensus} \cite{Fischler1981}.

Non randomized techniques, such as gradient ascent or descent are applicable, if gradients are available. In \cite{Le2021}, two deterministic methods using a non-smooth penalty method on the one hand, and a Alternating Direction Method of Multipliers (ADMM) on the other hand have been proposed. Those approaches solve convex subproblems to increase the optimization performance.
However, fundamentally, they can not guarantee global optimality either.

In contrast, exact methods are able to give guarantees, however, they are computationally costly since maximum consensus is fundamentally NP-hard, meaning it cannot be solved in polynomial time \cite{Chin2020}. Search strategies can be applied to limit the effort, such as \textit{Branch-and-Bound} (BnB) and tree search \cite{ChinSuter2017}.  
In \cite{Li2009}, \cite{Zheng2011}, \cite{Breuel1992}, \cite{Chin2014}, and \cite{Bustos2016}, BnB algorithms are proposed. In \cite{Chin2017}, a tree search approach is introduced and \cite{Olsson2008,Enqvist2012} use exhaustive search algorithms based on enumeration.
However, any such search strategies only provide a significant benefit if they are able to prune large portions of the search space, which is only the case if most of the solutions are concentrated within a small region.
\section{APPROACH}
\label{sec:approach}

We presume that a rough initial pose is available. In this work, it is obtained from a GNSS/IMU system, however, in general, its origin is not relevant.
Based on the initial pose, the car sensor scan is transformed into the global coordinate system, where it is matched to the globally georeferenced map point cloud.
Furthermore, the estimated parameters are restricted to the $xy$-position in the plane and the heading angle $\theta$, resulting in three degrees of freedom.

Although, in theory, the approach does not need initial values, we assume that the true position and heading are within a certain range around the initial pose and restrict the search range according to these assumptions. 
Of course, in practice, it is crucial that the search range is chosen large enough to reliably contain the true pose, e.~g.\ by including available uncertainty information of the initial pose.

To determine the pose within the defined search space, which is maximizing the robust objective function for the original as well as for the new maximum consensus approach, the search space is discretized and conveyed to a grid.
Then, the matches between scan and map points are identified for every cell. A match occurs, if a scan and a map point are within the $l_\infty$-distance defined by the edge length of a grid cell.


In the original maximum consensus approach, the simple count of matches is used for the consensus function $\Psi$, which is to be maximized to obtain the optimal rotation and translation parameters $\theta^*$ and $\boldsymbol{t}^*_{xy}$, formally
\begin{equation}\label{equ:1}
    \theta^*, \boldsymbol{t}^*_{xy} =
	\underset{\theta \in [-\pi,\pi),\: \boldsymbol{t}_{xy} \in \mathbb{R}^{2}}{\arg \max} \: \Psi(\boldsymbol{R}_\theta,\boldsymbol{t})
\end{equation}
\begin{equation}\label{equ:2}
	\Psi(\boldsymbol{R}_\theta,\boldsymbol{t}) = \sum_{i=1}^{n}\sum_{j=1}^{m}\mathbb{I}\bigl({\Vert\boldsymbol{R}_\theta{}\boldsymbol{p}_{i} + \boldsymbol{t} - \boldsymbol{q}_{j} \Vert}_{\infty} \leq \epsilon \bigr),
\end{equation}
where $\epsilon$ is the distance threshold and $\mathbb{I}(\cdot)$ is the indicator function.

\begin{figure}[ht!]
  \centering
  \includegraphics[width=0.7\columnwidth]{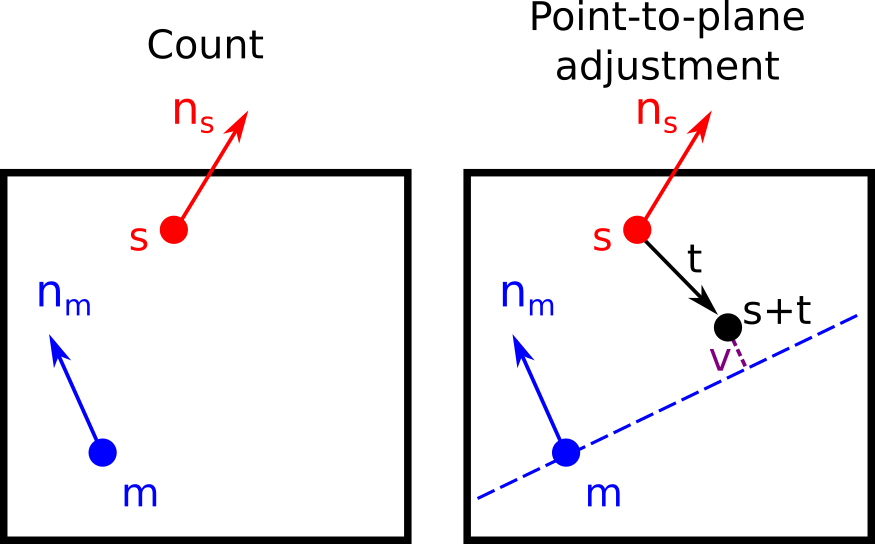}
  \caption{Difference between the original `count' and the new `point-to-plane adjustment' approach. For a matching scan and map point, the consensus function of the original approach is incremented by $+1$, whereas for the consensus function of the new approach, a point-to-plane observation equation is set up.} \label{fig:approach_comparison}
\end{figure}

Instead, for the new consensus function, a point-to-plane adjustment is carried out.
Figure \ref{fig:approach_comparison} shows the fundamental differences of both approaches. For a matching scan and map point,
the consensus function of the `count' approach is incremented by $+1$, whereas  
for the consensus function of the `point-to-plane adjustment' approach, the normal vector of the map point $\boldsymbol{n}_m$ is included in the evaluation over all cells.

Given that correspondences are between points and surfaces, the observation equation models the residual $v$ as the orthogonal distance,
\begin{equation}
\begin{aligned}\label{equ:3}
    0 + v &= \langle \boldsymbol{s} + \boldsymbol{t} - \boldsymbol{m} , \boldsymbol{n}_m \rangle \\
    \langle\boldsymbol{m} - \boldsymbol{s}, \boldsymbol{n}_m\rangle + v &= \langle \boldsymbol{n}_m, \boldsymbol{t}\rangle
= \boldsymbol{n}_m^\top \cdot \boldsymbol{t},
\end{aligned}
\end{equation}
where $\boldsymbol{s}$ is the scan point, $\boldsymbol{m}$ is the corresponding map point, $\langle\cdot,\cdot\rangle$ denotes the scalar product, and $\boldsymbol{t}$ is the translation to be estimated.

Setting up this observation equation for every match,  
the uncertainty of the unknown translation is described by the cofactor matrix $\boldsymbol{Q_{tt}}$, which is calculated based on the design matrix $\boldsymbol{A}$ and the weight matrix $\boldsymbol{P}$,
\begin{equation}\label{equ:4}
\begin{aligned}
\boldsymbol{Q_{tt}} = & \boldsymbol{N}^{-1} = (\boldsymbol{A}^\top \boldsymbol{P} \boldsymbol{A})^{-1}\\
= &\begin{bmatrix}
\sum{w_i n_{x,i}^2} & \sum{w_i n_{x,i}n_{y,i}}\\
\sum{w_i n_{x,i}n_{y,i}}& \sum{w_i n_{y,i}^2}
\end{bmatrix}^{-1}
\end{aligned}
\end{equation}
\begin{equation*}
\mbox{with}\; \boldsymbol{A} = \begin{bmatrix}
n_{x,1} & n_{y,1}\\
n_{x,2}& n_{y,2}\\
\vdots & \vdots \\
n_{x,k}& n_{y,k}\\
\end{bmatrix}\!\!,\; 
\boldsymbol{P} = \begin{bmatrix}
w_1& 0 & \dots&0\\
0& w_2& \dots&0\\
 \vdots&\vdots& \ddots& 0\\
0& 0&0&w_m\\
\end{bmatrix}\!\!,
\end{equation*}
%
%
%
where the design matrix $\boldsymbol{A}$ contains the normal vectors $\boldsymbol{n}_i$ of the matched map points. For the weight matrix $\boldsymbol{P}$, we use a diagonal matrix with weights defined by the cosine of the angle between $\boldsymbol{n}_{i,\text{scan}}$ and $\boldsymbol{n}_{i,\text{map}}$, as long as this is positive. Otherwise, the weight is zero, effectively disabling the correspondence. This is achieved by setting $w_i = \max(0, \langle \boldsymbol{n}_{i,\text{scan}}, \boldsymbol{n}_{i,\text{map}}\rangle)$.  

The cofactor matrix $\boldsymbol{Q_{tt}}$ indicates the relations between the covariances of the estimates. Using an eigendecomposition, $\boldsymbol{Q_{tt}} = \boldsymbol{U} \boldsymbol{\Lambda} \boldsymbol{U}^\top$, with $\boldsymbol{\Lambda} = \mbox{diag}(\lambda_1,\lambda_2)$, the eigenvalues $\lambda_1$ and $\lambda_2$ indicate (up to a factor) the minimum and maximum variance, in the respective (eigenvector) directions.
The observations lead to a precise determination of $\boldsymbol{t}$ if both, $\lambda_1$ and $\lambda_2$ are small, so that a suitable definition of a loss would be
$l = \max(\lambda_1, \lambda_2)$. Another definition, easier to handle, is $l' = \lambda_1 + \lambda_2$. For example, for 2D point unknowns, the square root of the sum of variances, $\sqrt{\sigma_x^2 + \sigma_y^2}$, is often used, also known as {\em Helmert's point error}. We therefore minimize (the square of) Helmert's point error, $\lambda_1 + \lambda_2 = \text{tr}(\boldsymbol{\Lambda}) = \text{tr}(\boldsymbol{Q_{tt}})$, or alternatively maximize
%
%
%
%
\begin{equation}\label{equ:6}
\mbox{score} = \frac{1}{\lambda_1+\lambda_2} = \frac{1}{\operatorname{tr}(\boldsymbol{Q_{tt}})}. 
\end{equation}

To see why this is a reasonable generalization of the simple consensus count score, observe that for $n_1$ observations of unit variance and unit weight in one (normal vector) direction, and $n_2$ such observations in orthogonal direction, $\text{score}=1/(1/n_1 + 1/n_2)$, so that e.~g.\ for $n_1=n_2=n$, $\text{score}=n/2$, growing linearly with $n$.
On the other hand, if $n_1=1$ (just one observation), then $\text{score}=n_2/(n_2+1)$, i.~e.\ $\text{score} < 1$, no matter how large $n_2$, reflecting the fact that the translation is not well determined. Using
$\operatorname{tr}(\boldsymbol{Q_{tt}}) =
\operatorname{tr}(\boldsymbol{N}^{-1}) = 
\operatorname{tr}(\boldsymbol{N}) / \operatorname{det}(\boldsymbol{N})$,
the score can be computed without having to use matrix inversion or eigendecomposition.

In this work, the search range around the initial pose is set to $6m \times 6m$ for the translation in the $xy$-plane (from $-3m$ to $+3m$ for both axes) and $\pm 2^{\circ}$ for the heading angle.
The edge length of a grid cell is set to $6cm$ resulting in $100 \times 100$ cells and the heading search range is evaluated using a step size of $0.5^{\circ}$, resulting in nine discrete angles.

\begin{figure}[ht]
  \centering
  \includegraphics[width=1\columnwidth]{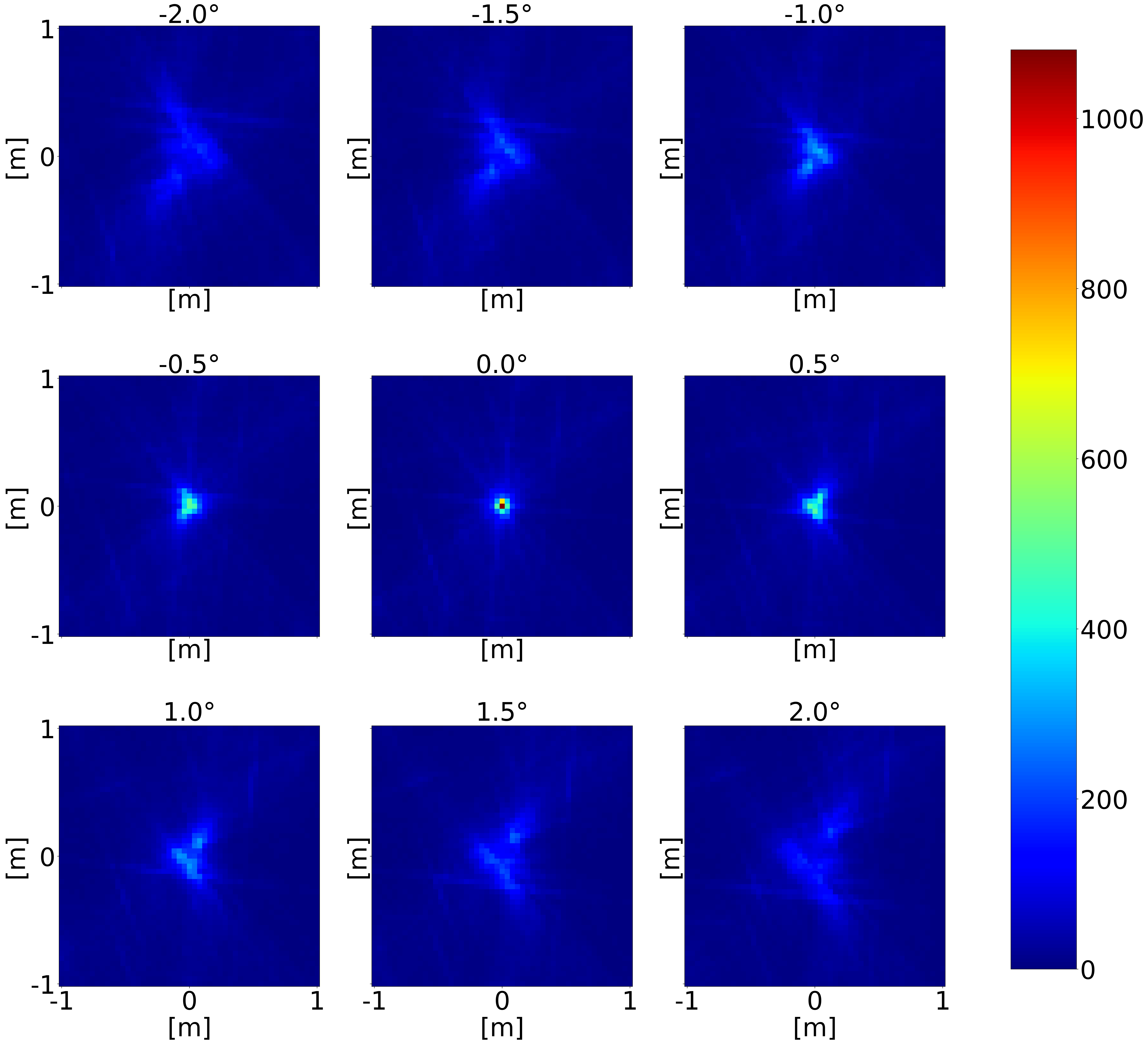}
  \caption{Search spaces of all nine investigated angles within the heading search range from $-2^{\circ}$ to $+2^{\circ}$ based on Helmert's point error (only section from $-1m$ to $+1m$ of the search space is depicted). The heading angle $\theta = 0^{\circ}$ results in the highest score.}\label{fig:heading_cov}
\end{figure}

Figure \ref{fig:heading_cov} shows the point-to-plane adjustment score, in terms of the search spaces corresponding to all nine angles within the heading search range from $-2^{\circ}$ to $+2^{\circ}$.
The highest count and score, respectively, occur at $\theta = 0^{\circ}$ and with increasing deviation from this heading angle, the consensus values decrease and the clear peaks vanish.
%
%

\begin{figure}[ht!]
  \centering
  \includegraphics[width=1\columnwidth] {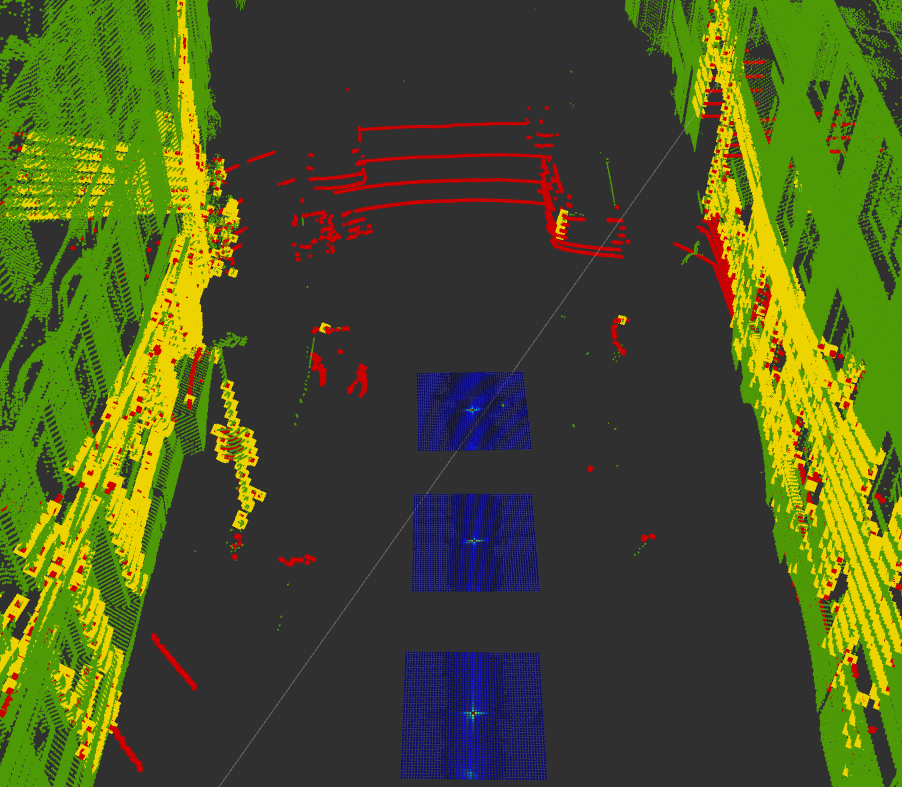}
 \caption{Maximum consensus localization based on the point-to-plane adjustment score. The dense and globally georeferenced map point cloud is visualized in green, a car sensor scan, after its transformation based on the initial pose, in red and the inlier set of the cell with the highest score in yellow. Each of the grids on the ground shows an evaluated 2D search space of a single epoch (colors use temperature scale, from blue to red). 
}\label{fig:approach}
\end{figure}

Figure \ref{fig:approach} shows the implemented maximum consensus localization based on the point-to-plane adjustment score. The dense and globally georeferenced map point cloud is visualized in green, the car sensor scan after the transformation based on the initial pose in red, and the inlier set of the cell with the highest score in yellow. 
In the map point cloud, ground points as well as parked cars have been removed beforehand, which is done based on semantic segmentation using simultaneously acquired images.
%
%
Consequently, car sensor scan points on the ground as well as on parked cars can not match any map points, do not add to the consensus set, and appear as outliers. In the figure, the grids on the ground visualize evaluated 2D search spaces for the heading, each of a single epoch, leading to the highest consensus value.
The red area within the search space indicates the position resulting in the highest consensus score and consequently, together with the heading, represents
the localization solution. For better visibility, only every third evaluated search space is visualized.

\begin{figure*}
  \centering
  \includegraphics[width=1\textwidth]{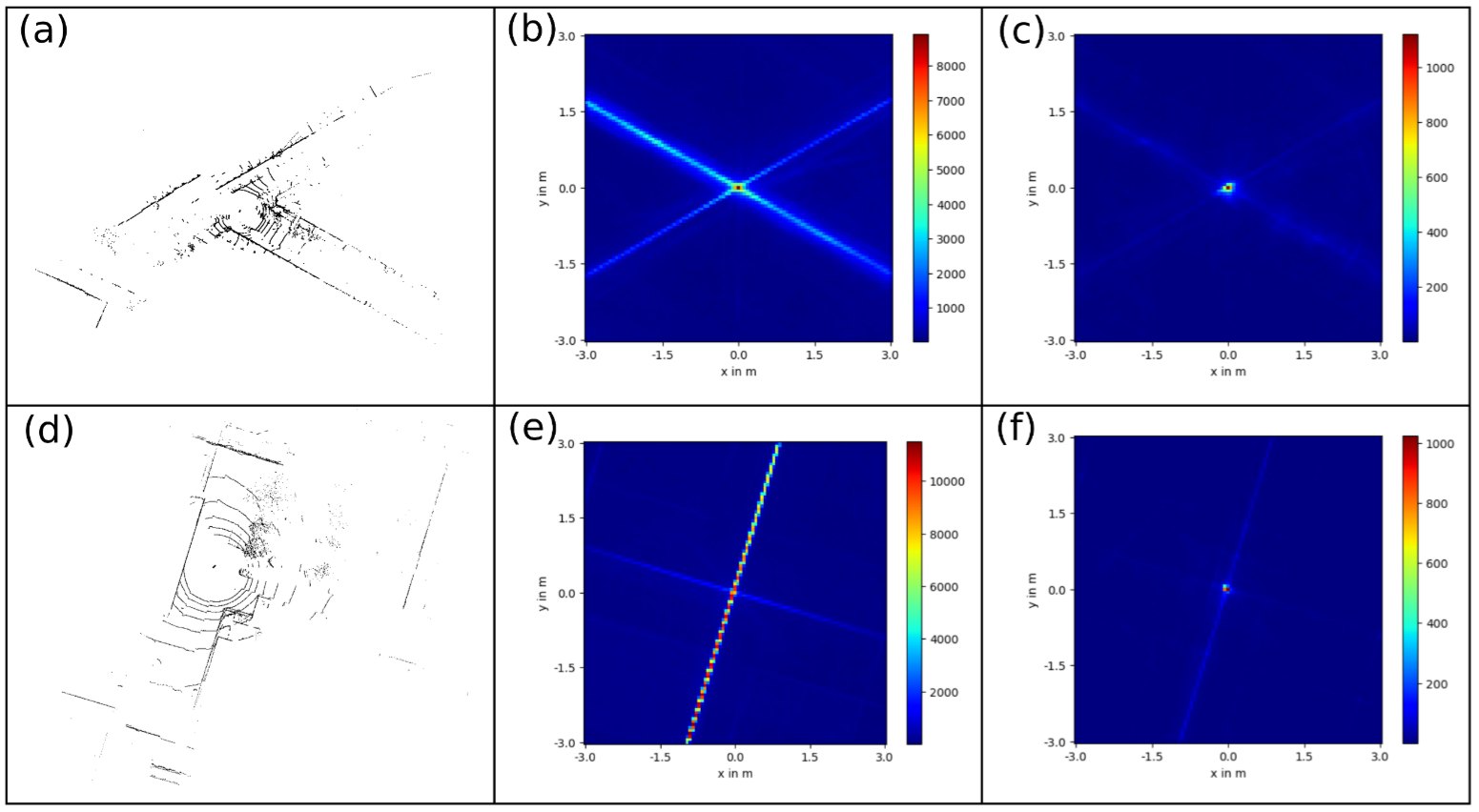} 
\caption{Image (a) shows a point cloud of a crossing scene, for which both the count (b) and the point-to-plane adjustment score (c) provide a distinct solution and find the correct position. In contrast, image (d) shows a point cloud, for which the highest consensus of the count (e) occurs at $(x=-0.9,y=-2.82)$, strongly deviating from the true solution, while the highest consensus of the point-to-plane adjustment score (f) occurs at $(x=0,y=0)$ matching the true solution.}
  \label{fig:merged}
\end{figure*}
%

Figure \ref{fig:merged} shows exemplarily the results of the original as well as new maximum consensus approach for a crossing situation (first row) and a situation, where the original approach fails (second row).
All four shown search spaces belong to the heading angle $\theta = 0^{\circ}$. Furthermore, 
each highest consensus is supposed to be at $(0,0)$. This is because the grids are centered in the known reference position, so that
if the maximum consensus localization works properly, 
the highest scores have to be at $(0,0)$.
With respect to the point cloud of the crossing scene (a), this is the case for the count (b) and the point-to-plane adjustment score (c), which both provide a distinct solution and find the correct position. 
In contrast, image (d) shows a point cloud, for which the highest consensus of the count (e) occurs at $(x=-0.9,y=-2.82)$ strongly deviating from the true solution, and the highest consensus of the point-to-plane adjustment score (f) occurs at $(x=0,y=0)$ matching the true solution. 
Those failures of the count can occur due to an irregular point density within the map.  
Since the car sensor scans have an irregular point density and the objective function aims for a maximization of matches, the approach inherently 
overlaps areas with high point densities.
So whenever there is an irregular point density in the map, the count objective function is susceptible to fail.  

In general, with respect to the original objective function, for a straight street, one strong ray containing multiple high consensus values appears, and for a crossing, two rays arise intersecting in one sharp and unique peak.
Whereas for a straight street without any side roads, parallel facades on both sides are the dominant structure, rendering perpendicular structures invisible, for a crossing, facades with normal vectors pointing in different directions unambiguously constrain the pose leading to a clear and unique solution.
Single planes inherently constrain possible poses to a parallel shift along each plane, which results in rays of high consensus values, which in turn, if multiple rays exist, intersect in one point: the localization solution. 
However, for a straight street situation, the certainty of the localization solution is low due to multiple grid cells along the single ray yielding high consensus values.

In contrast to the original maximum consensus approach, the new approach also provides a unique localization solution for straight streets. As intended, the score of consensus sets only with normal vectors pointing in one direction is reduced and of consensus sets with normal vectors in various directions is high.
This reduces the susceptibility to failures arising from irregular point densities tremendously.
For a crossing scene, the rays of the original approach, which are leading to the clear peak at their intersection, vanish, and only the unique peak remains, eventually even more distinct. 

\section{DATA}
For the measurement campaign, a circular trajectory in an inner city area characterized by a dense building structure, narrow streets and many parked cars was driven five times. 
%
The car sensor scans were acquired using a Velodyne VLP-16 \cite{Velodyne2016}, which has 16 scan layers in a vertical field of view from $-15^\circ$ to $+15^\circ$ with a uniform spacing of $2^\circ$ between two layers.
The horizontal field of view is 360° with a horizontal spacing of $0.2^\circ$ between two consecutive points. The accuracy of this sensor is $\pm3cm$.

The dense and highly accurate point clouds as well as the trajectory, on which the map data is based, were acquired by a RIEGL VMX-250 Mobile Mapping System (MMS) \cite{Riegl2012}. Specifically, the point clouds were gathered by two built-in RIEGL VQ-250 LiDAR sensors,
having a ranging accuracy of ten millimeters. The trajectory was obtained using an Applanix POS LV GNSS/IMU system and corrected using post-processing with correction data from the Satellite Positioning Service (SAPOS).
%
The point clouds and trajectories from the five rounds of the measurement campaign were aligned using the adjustment process from Brenner \cite{Brenner2016}. Afterwards, the standard deviation of the map is below two centimeters. 
%
Since the map point cloud and the car sensor scans were acquired during the same measurement run, temporary static objects, in particular parked cars, need to be removed from the map, as they would cause a too optimistic localization result. Also, for a general use case, those objects should be removed since they would always influence the localization. Apart from parked cars, points on the ground and on vegetation were removed. This was done using a semantic segmentation similar to \cite{petersEA2020}.

\section{EVALUATION}
The maximum consensus localization both based on the count and the point-to-plane adjustment score is conducted for all 3001 epochs of one measurement round.
Additionally, a map matching based on ICP is computed for every epoch. The ICP is initialized using a $5\times5$ grid, centered at the ground truth position, with a grid spacing of $1m$. For each of the resulting 25 ICP runs per epoch, it is evaluated to where the ICP converges.

The evaluation consists of two parts.
First, the maximum consensus localization is compared to the ICP result. 
Second, the maximum consensus approaches based on the count and point-to-plane adjustment score are compared to each other considering three different criteria with respect to the distribution within the search space: the ratio between the highest and second highest consensus, 
the kurtosis along the significant ray, 
and the Kullback-Leibler divergence with respect to a Laplace distribution.

The maximum consensus localization corresponds to a grid based search, whereas the ICP, apart from the step of finding correspondences, consists of a least squares estimation.
Consequently, if no sub-pixel interpolation is applied, the accuracy of the maximum consensus approach is limited to the resolution of the search space, while the ICP - given the correct correspondences - finds the exact solution. 
However, in this work, the focus is not on which approach provides the higher accuracy of the final localization, but it is on the ability to find the correct solution at all - independent of a final adjustment.
In this regard, the maximum consensus localization only requires that the true solution is within its search radius, whereas the ICP requires an initialization which goes beyond that. Our evaluation based on the grid-wise initialization of the ICP confirms this shortcoming. Whereas the maximum consensus localization based on the point-to-plane adjustment score always finds the true pose, the ICP fails in more than 20 percent of the epochs. Here, failing corresponds to the case that not all ICP initializations converge to the true solution.

\begin{figure*}%
    \centering
    \subfloat[
    Ratio between the highest and second highest consensus: The point-to-plane adjustment score leads to a decrease of the second highest consensus values indicating more distinct localization solutions.]{{\includegraphics[width=0.6\columnwidth]{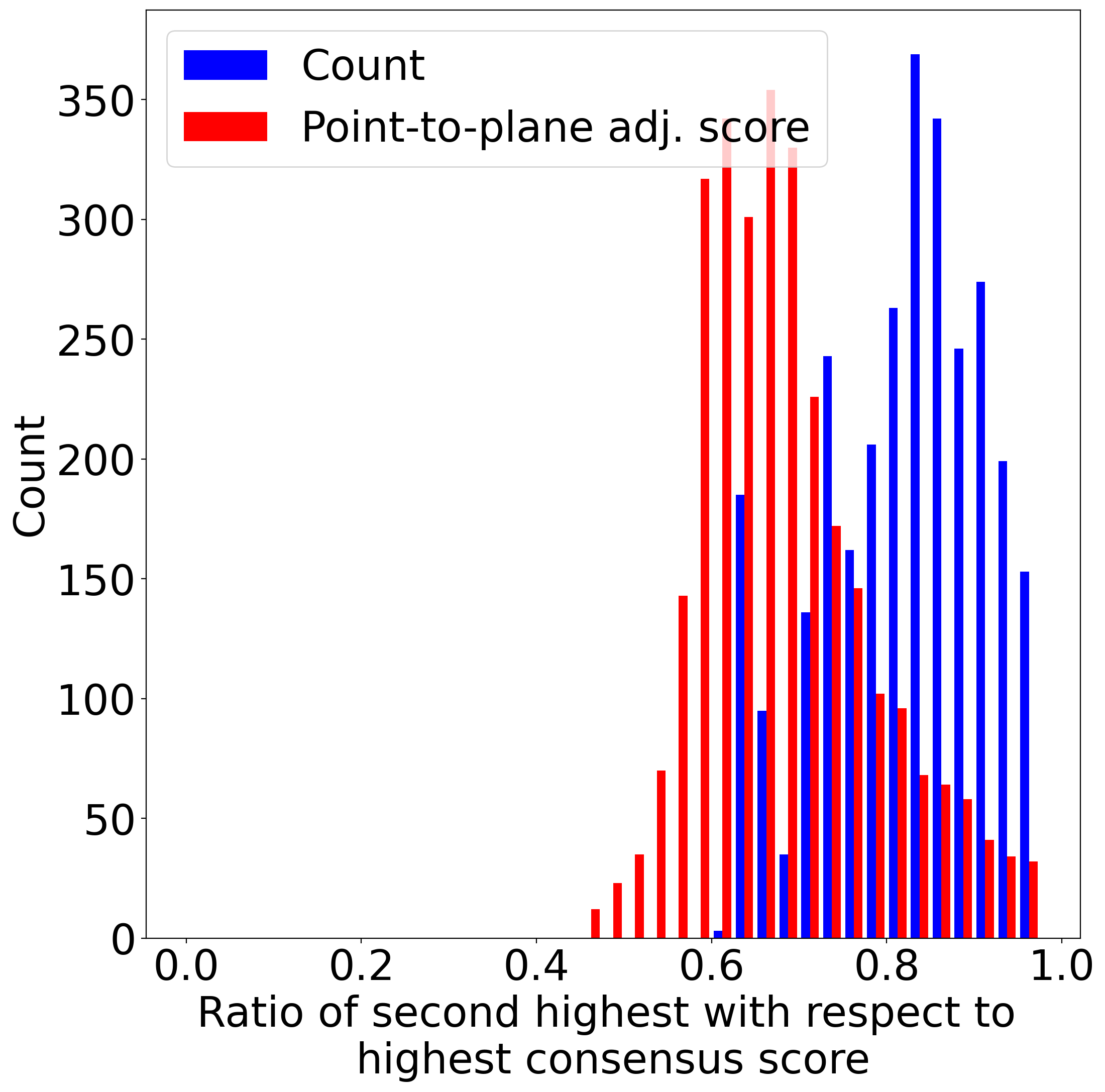} }}%
    \qquad
    \subfloat[
    Kurtosis along the significant ray within the search space: The point-to-plane adjustment score leads to an increase of the kurtosis indicating more distinct localization solutions.]{{\includegraphics[width=0.6\columnwidth]{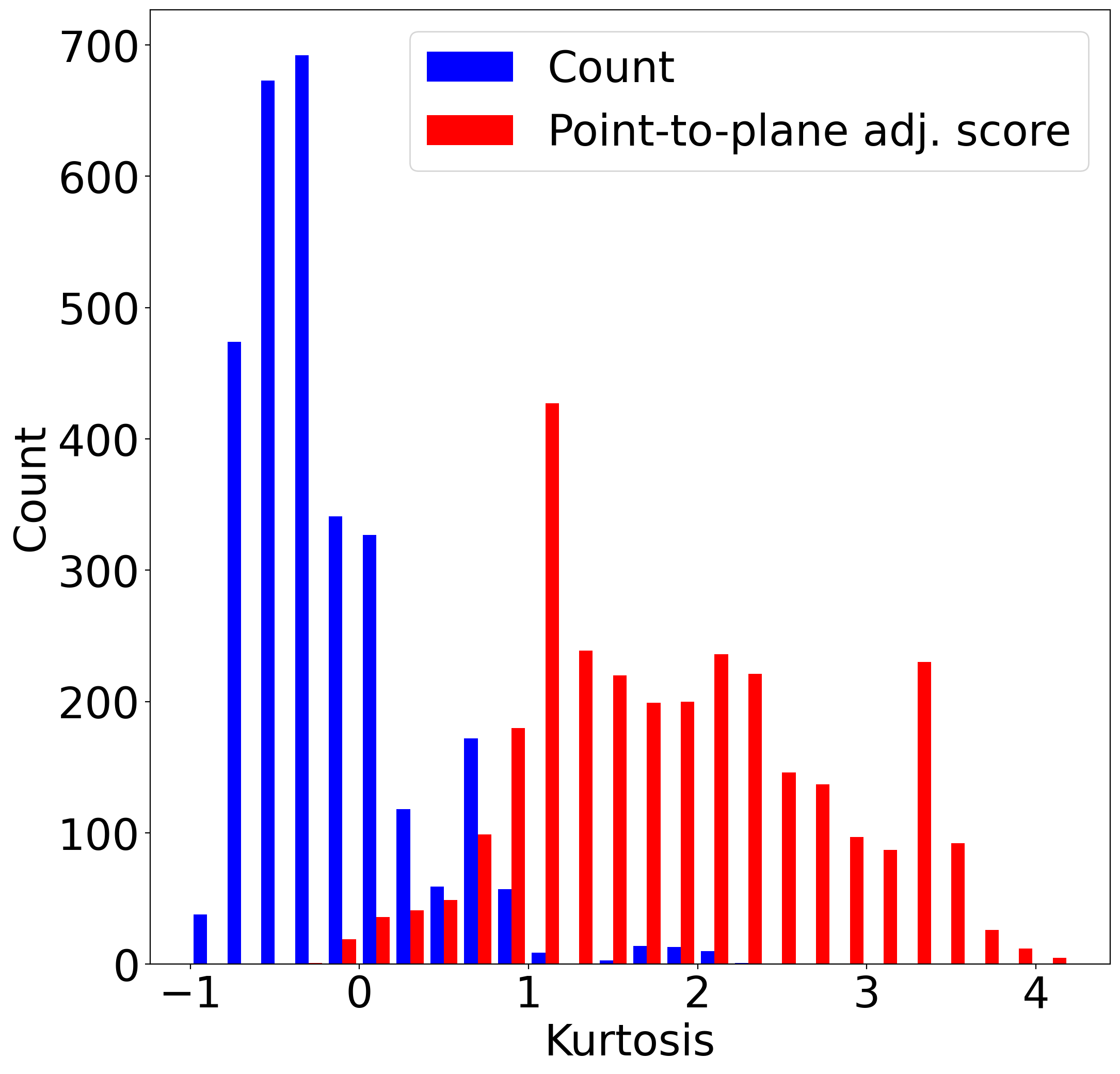} }}%
    \qquad
    \subfloat[
    Kullback-Leibler divergence with respect to a Laplace distribution: The point-to-plane adjustment score leads to a reduction of the Kullback-Leibler divergence indicating more distinct localization solutions.]{{\includegraphics[width=0.6\columnwidth]{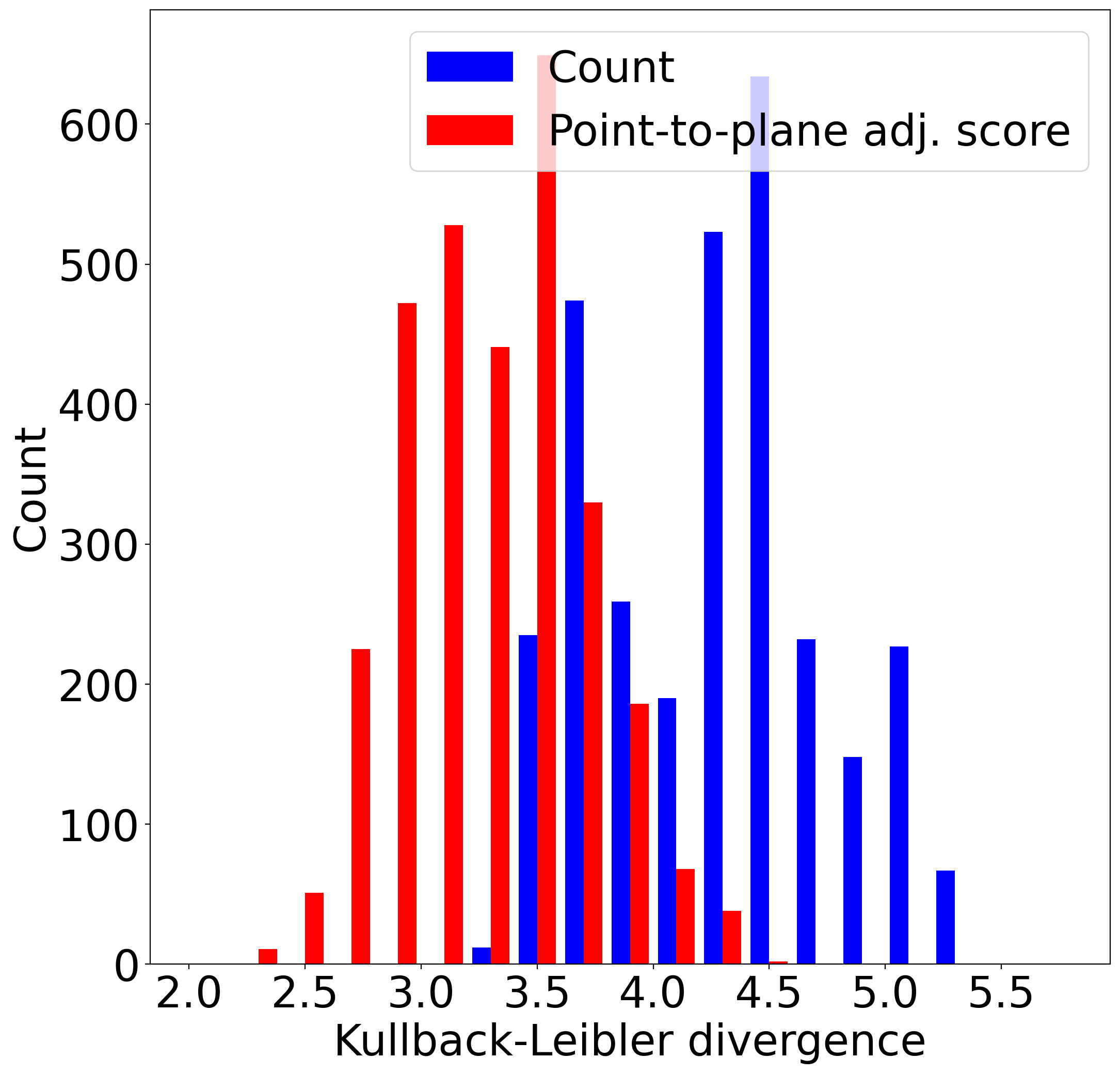} }}%
   
    \caption{Comparisons between maximum consensus approaches based on the count (blue) and point-to-plane adjustment score (red) over all 3001 epochs of the measurement run.}%
    \label{fig:evaluations}%
\end{figure*}

As we have described in Section \ref{sec:approach} and shown in Figure \ref{fig:merged}, in the presence of irregular point densities, the basic count of matches as objective function may fail.
In addition, apart from the worst case that the localization fails, there are many epochs where the count finds the correct pose, however, while being close to fail. 
To examine how distinct the solutions are, for both objective functions, in every epoch, the ratio between the highest and second highest consensus score, the kurtosis along the significant ray, and the Kullback-Leibler divergence with respect to a Laplace distribution are computed and considered as measures of certainty. The results are depicted in Figure \ref{fig:evaluations}. 
With respect to the ratio between the highest and second highest consensus shown in Figure \ref{fig:evaluations} (a), compared to the count, the point-to-plane adjustment score yields an overall reduction of the second highest consensus values and consequently more distinct and reliable localization solutions. However, some critical epochs remain. For those, an additional investigation is conducted. The largest distance between the grid cell of the highest consensus and a grid cell reaching at least 90\% of the highest consensus is considered. The results show that for the point-to-plane adjustment score, always only neighboring cells reach high consensus scores, whereas for the count, high consensus values arise within the entire search space, which confirms the less critical and more reliable characteristic of the point-to-plane adjustment score compared to the count.
With respect to the kurtosis along the significant ray within the search space shown in Figure \ref{fig:evaluations} (b), the point-to-plane adjustment score leads to a strong increase. The kurtosis describes the \textit{tailedness} of a distribution. Since a low kurtosis corresponds to a flat, rather uniform distribution and a high kurtosis indicates a distinct peak, 
this measure shows the tremendous improvement with respect to an unique registration and a reliable localization.
Finally, the Kullback-Leibler divergence with respect to a Laplace distribution with $\sigma = 1 cell$ is considered. The Laplace distribution is characterized by a significant peak similar to the one we are claiming for a distinct and reliable solution. 
Considering the result shown in Figure \ref{fig:evaluations} (c), the point-to-plane adjustment score yields a reduction of the Kullback-Leibler divergence indicating a better fit with respect to the Laplace distribution and confirming the conclusion of the peak ratio as well as the kurtosis that the new objective functions provides a more reliable registration and, for our purpose, a more reliable localization.

\begin{figure}[ht!]
  \centering
  \includegraphics[width=1\columnwidth]{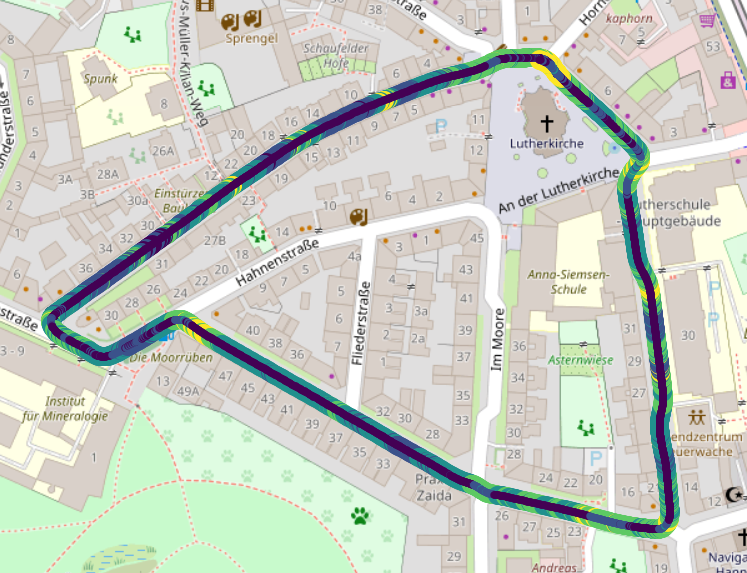}
  \caption{Trajectory of the measurement campaign visualizing the kurtosis values for the count (small circles) and point-to-plane adjustment score (large circles). Dark blue indicates a low kurtosis, yellow indicates a high kurtosis (background map: OpenStreetMap \cite{OpenStreetMap}).}\label{fig:kurtosis_count_score}
\end{figure}

Exemplarily, in Figure \ref{fig:kurtosis_count_score}, the kurtosis 
of the accumulator is shown for the count (small circles) and the point-to-plane adjustment score (large circles) over the measurement trajectory. Dark blue indicates a low kurtosis, while yellow indicates a high kurtosis. In accordance with the histogram of Figure \ref{fig:evaluations} (b), the kurtosis values of the count are basically lower. Only at crossings, the kurtosis of the count increases, however, still being low in relation to the point-to-plane adjustment score, which provides comparatively high kurtosis values in straight streets and very high kurtosis values at crossings.

\section{CONCLUSIONS}
In this paper, we have introduced a novel robust objective function for maximum consensus point cloud registration, particularly applied on scans from a Velodyne VLP-16 and a highly accurate, dense map point cloud for the purpose of vehicle localization.
The objective function is based on Helmert's point error, which is a scalar measure describing the accuracy 
of a least squares adjustment.
The new approach successfully tackles the shortcomings of the common maximum consensus strategy of simply counting matches. 
In general, the introduced objective function increases the reliability of the registration and hence the localization solution. It allows vehicle localization in the presence of strongly outlier contaminated data as well as challenging environments. 
The new maximum consensus localization approach can be incorporated in a filter integrating previous epochs and other sensors 
as well as enhanced to a SLAM approach including also temporary static objects 
in the map, which were excluded in the present paper.

\addtolength{\textheight}{-0cm}   





\section*{ACKNOWLEDGEMENT}
This  project  is  supported  by  the  German  Research  Foundation  (DFG),  as  part of the  Research  Training  Group  i.c.sens, GRK 2159, `Integrity and Collaboration in Dynamic Sensor Networks'. 


\end{document}